\documentclass[conference]{IEEEtran}
\usepackage{cite}

\ifCLASSINFOpdf
   \usepackage[pdftex]{graphicx}
   \graphicspath{{./figs/}}
   \DeclareGraphicsExtensions{.pdf,.PDF,.jpeg,.JPG,.png,.eps}
\else
\fi
\usepackage{amsmath}
\usepackage{amssymb}
\usepackage{amsfonts}
\usepackage{amsthm}
\usepackage{manfnt}
\usepackage{mathtools} 
\usepackage{array}
\usepackage{booktabs}

\usepackage{array}
\usepackage{tabularx, booktabs}
\usepackage{multirow}

\ifCLASSOPTIONcompsoc
  \usepackage[caption=false,font=normalsize,labelfont=sf,textfont=sf]{subfig}
\else
  \usepackage[caption=false,font=footnotesize]{subfig}
\fi
 \usepackage{dblfloatfix}
 \usepackage{tabularx}
 \usepackage[margin=2.5cm]{geometry}
\usepackage{multirow,tabularx}

\usepackage{color}

\usepackage{url}

\begin{document}
\title{Convolutional Drift Networks\\for Video Classification}

\author{\IEEEauthorblockN{Dillon Graham\IEEEauthorrefmark{4},
Seyed Hamed Fatemi Langroudi\IEEEauthorrefmark{4},Christopher Kanan\IEEEauthorrefmark{2} and
Dhireesha Kudithipudi\IEEEauthorrefmark{4}}
\IEEEauthorblockA{ \IEEEauthorrefmark{4}NanoComputing Research Lab, \IEEEauthorrefmark{2}kLab\\ Rochester Institute of Technology\\
Rochester, New York 14623\\}}




\maketitle

\begin{abstract}
Analyzing spatio-temporal data like video is a challenging task that requires processing visual and temporal information effectively. Convolutional Neural Networks have shown promise as baseline fixed feature extractors through transfer learning, a technique that helps minimize the training cost on visual information. Temporal information is often handled using hand-crafted features or Recurrent Neural Networks, but this can be overly specific or prohibitively complex. Building a fully trainable system that can efficiently analyze spatio-temporal data without hand-crafted features or complex training is an open challenge. We present a new neural network architecture to address this challenge, the Convolutional Drift Network (CDN). Our CDN architecture combines the visual feature extraction power of deep Convolutional Neural Networks with the intrinsically efficient temporal processing provided by Reservoir Computing. In this introductory paper on the CDN, we provide a very simple baseline implementation tested on two egocentric (first-person) video activity datasets. We achieve video-level activity classification results on-par with state-of-the art methods. Notably, performance on this complex spatio-temporal task was produced by only training a \emph{single} feed-forward layer in the CDN.



\end{abstract}
%
\begin{IEEEkeywords} 
Deep Learning, Reservoir Computing, Video Activity Classification
\end{IEEEkeywords}

\IEEEpeerreviewmaketitle

\section{Introduction}
\IEEEPARstart{D}{eep} neural networks have significantly advanced the state-of-the-art in computer vision~\cite{ResNet50}, natural language processing~\cite{Xia2016}, speech recognition~\cite{Amodei2015}, and  robotics~\cite{Lenz2015}. These networks are very effective at extracting high-level, complex abstractions of input data through a hierarchical learning process. Deep Convolutional Neural Networks (CNNs) achieve superior performance in visual object recognition tasks, and they have largely replaced hand-crafted features as the standard approach in this area. While deep learning is advantageous for large amounts of spatial data, it also has limitations. Using these networks for temporal data (e.g. video analysis) introduces several new challenges, typically addressed using Recurrent Neural Networks (RNNs). Modern RNN models like Long Short-Term Memory (LSTM) are an effective way to handle temporal data, but they also tend to be difficult and expensive to train. In video analysis, LSTMs are often paired with CNNs, but this is likely to increase network and training complexity for most tasks. A simpler method for analyzing spatio-temporal data is desirable, and video analysis tasks are complex enough to reasonably test new methods. 

Video is now a ubiquitous source of spatio-temporal data, and interest in video analysis has risen due to the increasing presence of video data on the internet. Despite the rise of video data availability, video analysis remains a relatively under-examined area compared to image analysis. Many approaches still focus on hand-crafted features akin to those historically used in computer vision. Works applying deep learning to this domain with CNNs and hierarchical layers of LSTMs have shown results \cite{Ng2015}, but combining a CNN with hierarchical layers of LSTMs necessitates training a very large number of parameters, tuning many different hyper-parameters, and performing backpropagation through time. These requirements can be prohibitive, especially in real-world applications with size, weight, area, and power constraints (e.g. robotics, remote sensing, autonomous vehicles). 

Visual information in video can be processed on a per-frame basis using CNNs, but training a randomly initialized deep network capable of operating on high resolution image data requires large amounts of \emph{labeled} data. Without sufficient training data, these networks are prone to overfitting. Unfortunately, many datasets do not contain enough labeled training samples for networks to converge effectively. In computer vision, this problem has been partially resolved using transfer learning to improve performance and decrease training time. This is usually achieved by initializing a new network with layers and trained weights drawn from a publicly available high-performance CNN model. In most cases, these source networks were built to achieve state-of-the-art results for ImageNet, a dataset that contains 1000 different object categories over approximately one million labeled images. When trained on this kind of large natural image dataset, early convolutional layers in a CNN produce features with a surprising level of generality (i.e., features remain useful for most images) \cite{Razavian2014, Yosinski2014}. This feature generality is the primary characteristic which makes transfer learning with pre-trained CNNs so effective. Less work has been done in applying this kind of transfer learning to video datasets, and the best way to do this is still an open question.

RNNs can be used for processing temporal information in video, but successes are limited to LSTM and its variants. RNNs are often are either inherently complex, or have other limitations (e.g., capacity in Hopfield networks). Reservoir Computing (RC) offers some under-examined alternative RNN models, which are simple to construct and require minimal training. RC models are based on a neuroscientific model of corticostriatal processing loops~\cite{Lukosevicius2012}, and they have been studied in many different applications \cite{Verstraeten2006, Kudithipudi2015} as an unconventional RNN. Two well known RC models are the \textit{Liquid State Machine} (LSM)~\cite{Maass2010} and the \textit{Echo State Network} (ESN)~\cite{Jaeger2004a}. 

In this paper, we propose a novel neural network architecture, the Convolutional Drift Network (CDN). CDNs produce per-frame appearance features from video using a deep CNN, and those features are pushed into a randomly initialized ESN reservoir. Since the ESN reservoir topology and weights are static after initialization, features produced by the CNN are the driving force in this network. Essentially, once a feature is produced and pushed into the ESN, it is left to propagate naturally, or `drift' through the fading memory representation produced by the Echo State property of ESNs. We evaluate our model on two video classification datasets.  We also investigate how far we can simplify CDN training while still achieving competitive performance on this non-trivial classification task. Minimizing training complexity and resources is a step toward the eventual implementation of similar architectures on hardware for field-deployable systems. 
The specific contributions of this research are:
\begin{enumerate}
	\item We introduce a new neural network architecture for spatio-temporal tasks, the Convolutional Drift Network (CDN). 
	\item We bridge the gap between deep CNNs and ESNs. 
    \item We demonstrate that CDNs are competitive at video classification tasks with state-of-the-art methods, using only a \emph{single} trained neural network layer. \\
\end{enumerate} 

\section{Related Works}

%
%
%


\subsection{Video Activity Classification}
Video activity classification requires identifying appearance features present in video frames, often called feature descriptors. Feature descriptors are used to find correlated frames in a sequence. This has been accomplished with several methods, for example: creating bag-of-words histograms of CNN features ~\cite{Ryoo2015} and using LSTMs to process sequential CNN features ~\cite{Ng2015}. LSTMs outperform simpler temporal pooling techniques ~\cite{Ng2015}. However, training an LSTM requires using the backpropagation-through-time algorithm, which is expensive, both in time and energy. For video classification problems, this often requires multiple days, even when GPUs are used.

Video frames are not always strongly representative of a target activity. Correctly selecting frames indicative of a particular activity is a non-trivial task ~\cite{Ng2015}. Most techniques use pooling methods to aggregate all information about a video. Piergiovanni et al. \cite{Piergiovanni2016a} noted that pooling across an entire video produces classification mistakes for videos containing multiple events, and the authors proposed temporal attention filters as a potential remedy. Their approach focused attention temporally on the portion of a video that relates to specific activity.  

Egocentric videos are becoming increasingly pervasive (e.g., lifeloggers, police body cameras, and robotic platforms ~\cite{summarizationegocentric}). Video activity classification is generally more challenging in egocentric videos, which are recorded from a first-person perspective using a wearable device. Research in this emerging domain mostly focuses on learning after a video is acquired. However, on-device learning may be useful for applications where online learning or near real-time learning is desirable. RNNs that use backpropagation to train (e.g., LSTM) are heavily memory bound and are compute intensive. They are likely to be unsuitable for deployment on embedded/wearable devices with strict constraints on energy and compute resources.  

Traditional techniques used in video activity detection represent feature descriptors using Histogram of Oriented Gradients  or Histogram of Optical Flows and then combine these features using Bag of Visual Words  or Fisher Vector encodings  \cite{Ryoo2015,Herath2016}. Recently, neural network based video activity classification methods have been explored using 2D DCNNs or 3D DCNNs \cite{Tran2016}. Other methods use a composite network created by combining a pre-trained or untrained DCNN with other neural network elements to extract spatio-temporal information. In these composite networks DCNN layers are combined with layers containing recurrent neural networks, usually LSTM layers\cite{Herath2016,Baccouche2011}. During exploration of composite neural networks, many authors also chose to explore replacing or combining RNN elements with more traditional pooling techniques. This is a common type of solution that uses both neural networks and traditional hand-crafted feature techniques. For instance, Ryoo et al. proposed a new technique for feature pooling over time with first person video activity detection. This technique is intended to identify general features representative of each video, based on time-series analysis of frames \cite{Ryoo2015}.  Ng et al. consider temporal pooling techniques such as average pooling and max pooling applied to appearance features sampled from each video frame \cite{Ng2015} while also examining composite pre-trained DCNN and LSTM networks. Finally, Wang et al. combined handcrafted feature descriptors and DCNN appearance features to implement a pooling technique intended to improve video activity detection \cite{Wang2015}. 

Most of the video activity detection techniques mentioned above employ hand-crafted features or LSTMs. Hand-crafted features must be designed for a task by a subject matter expert. LSTM training can be complex, and time consuming. Neither methods are well suited to applications where a general and efficiently trainable system is desired. 



\section{Background}
\subsection{Transfer Learning with Deep CNNs}
Architectures like ResNet \cite{ResNet50} are now commonly used as pre-trained CNN models for many image analysis tasks. There are several types of transfer learning. In most cases, CNN networks trained on the ImageNet Large Scale Visual Recognition \cite{ImageNet2015} dataset are used. These source networks are useful for extracting appearance features from a given dataset, with better results than training with no pre-training. Since these source models contain stable weights learned during their original training, they offer a starting point to expedite training on a new dataset. Alternatively, pre-trained CNNs can be used to provide off-the-shelf CNN features \cite{Razavian2014} as static appearance feature extractors. Typically, a CNN source network is truncated at some layer, depending on the level of feature specificity desired. 


\subsection{Reservoir Computing}
Among RC models, ESNs are closest to contemporary neural networks which use conventional floating point values to represent weights. The reservoir in an ESN is a recurrent network, constructed by randomly generating synaptic weights and topology. Recurrent connections in the reservoir create a fading memory of an input, which can be used to represent sequences. Weights within the reservoir stay fixed after initialization (i.e. they are not trainable). The fixed nature of reservoir weights greatly simplifies the training process, avoiding costly error propagation through time. 

\subsection{Echo State Networks} \label{sec:bg_esn}
The ESN is the primary RC model of interest in this paper. 
A basic ESN operates in a very similar manner to other neural networks, with an input feature vector $\mathbf{u}(n) \in \mathbb{R}^{N_u}$, and an output vector $\mathbf{y}(n) \in \mathbb{R}^{N_y}$. The goal is to train a network to produce an output $\mathbf{y}(n)$ that best approximates some desired signal, $\mathbf{y}^{target}(n) \in \mathbb{R}^{N_y}$. As with any RNN model, time must be considered, and $n=1,\dots,T$ represents discrete time steps over data with $T$ sequential elements (e.g. video with 60 frames $\rightarrow T=60$). Activity within the reservoir is modeled with $\mathbf{x}(n) \in \mathbb{R}^{N_x}$, calculating some activation function at time $n$, and an update function $\mathbf{\tilde{x}}(n)$. These functions are usually a hyperbolic tangent activation function as shown in Eqns. \ref{equ:basic_res_activation_1}, \ref{equ:basic_res_activation_2}. 
\begin{align}
\mathbf{\tilde{x}}(n) &= tanh( \mathbf{W}^{in} [1;\mathbf{u}(n)] + \mathbf{W}^{x}\mathbf{x}(n-1) ) \label{equ:basic_res_activation_1}\\
\mathbf{x}(n) &= ( 1 - \alpha )  \mathbf{x}(n-1) + \alpha \mathbf{\tilde{x}}(n)
\label{equ:basic_res_activation_2}
\end{align}
\noindent
where, $\alpha \in (0,1]$ is a leaking rate coefficient, $tanh(\cdot)$ is computed element-wise, and $[ \cdot ; \cdot ]$ is vector concatenation.


In an ESN, three weight matrices define connections between the input layer, reservoir layer, and output layer. The input weights $\mathbf{W}^{in}  \in \mathbb{R}^{N_{x} \times (1+N_{u})}$ connect $\mathbf{u}(n)$ to the reservoir. Output weights $\mathbf{W}^{out} \in \mathbb{R}^{N_{y} \times (1 + N_u + N_{x})}$ connect the reservoir to the output $\mathbf{y}(n)$. Recurrent connections in the reservoir are represented by $\mathbf{W}^{x} \in \mathbb{R}^{N_{x} \times N_{x}}$. Some ESN variants use additional weights (e.g. weighted feedback connections), but a basic ESN can be constructed with these three matrices. Weights are all randomly initialized, but with additional constraints placed on $\mathbf{W}^{x}$ to control properties like connection sparsity, and non-zero weight distribution. 

Since all weights except $\mathbf{W}^{out}$ are left fixed during learning, network outputs are essentially feed-forward and linear. 
Learning is accomplished by simply optimizing $\mathbf{y}(n)$ to match $\mathbf{y}^{target}(n)$ by iteratively adjusting $\mathbf{W}^{out}$ with traditional methods like ridge regression. When performing sequence  classification, the common approach is to set $N_y=k$, where $k$ is the number of classes (i.e. $\mathbf{y}(n) \in \mathbb{R}^{k}$). In this configuration each $\mathbf{y}(n)$ output value $y_{1}, \dots, y_{k}$ corresponds to a single class. Finding the maximum activation for an input sequence $\mathbf{u}(n)$  provides the class label prediction, as shown in Eqns. \ref{equ:esn_basic_clf_1} and \ref{equ:esn_basic_clf_2}.
\begin{align}
class(\mathbf{u}(n)) &= \underset{k}{\arg\max} \left( \frac{1}{|\tau|} \sum_{n \in \tau} y_k (n) \right) \label{equ:esn_basic_clf_1}\\
&= \underset{k}{\arg\max} ( ( \Sigma \mathbf{y} )_k )
\label{equ:esn_basic_clf_2}
\end{align}
\noindent
where, $y_k (n)$ is the $k$th dimension of $\mathbf{y}(n)$ and $\Sigma \mathbf{y}$ is $\mathbf{y}(n)$ time-averaged over $\tau$. This value can easily be computed by multiplying $\mathbf{W}^{out}$ with $\mathbf{x}(n)$ activations time-averaged in a similar manner, as shown in Eqns. \ref{equ:esn_basic_clf_3} and \ref{equ:esn_basic_clf_4}.
\begin{align}
\Sigma \mathbf{y} &= \mathbf{W}^{out} \frac{1}{|\tau|} \sum_{n \in \tau} [ 1 ; \mathbf{u}(n) ; \mathbf{x}(n) ]
 \label{equ:esn_basic_clf_3}\\
	 &= \mathbf{W}^{out} \Sigma \mathbf{x}
\label{equ:esn_basic_clf_4}
\end{align}
\noindent
where, $\Sigma \mathbf{x}$ is $[ 1 ; \mathbf{u}(n) ; \mathbf{x}(n) ]$ time-averaged over $\tau$. This method computes $\Sigma \mathbf{y}$ through only one multiplication with $\mathbf{W}^{out}$. The value $\tau$ can represent an entire sequence length, or some interval within the sequence. Different $\tau$ values can also be used for variable length sequences. \\

Two hyper-parameters provide a large contribution to the performance of an ESN: the number of neurons $N_x$, and the spectral radius $\rho (\mathbf{W^x})$. The $N_x$ value tends to have the greatest performance impact, and it is generally thought that bigger reservoirs are better for challenging tasks. Spectral radius $\rho (\mathbf{W^x})$ helps determine how long an input impacts reservoir activations. A greater $\rho$ value is usually appropriate for tasks requiring a longer memory of input values.

\begin{figure*}[ht!]
	\begin{center}
		\includegraphics[width=\linewidth]{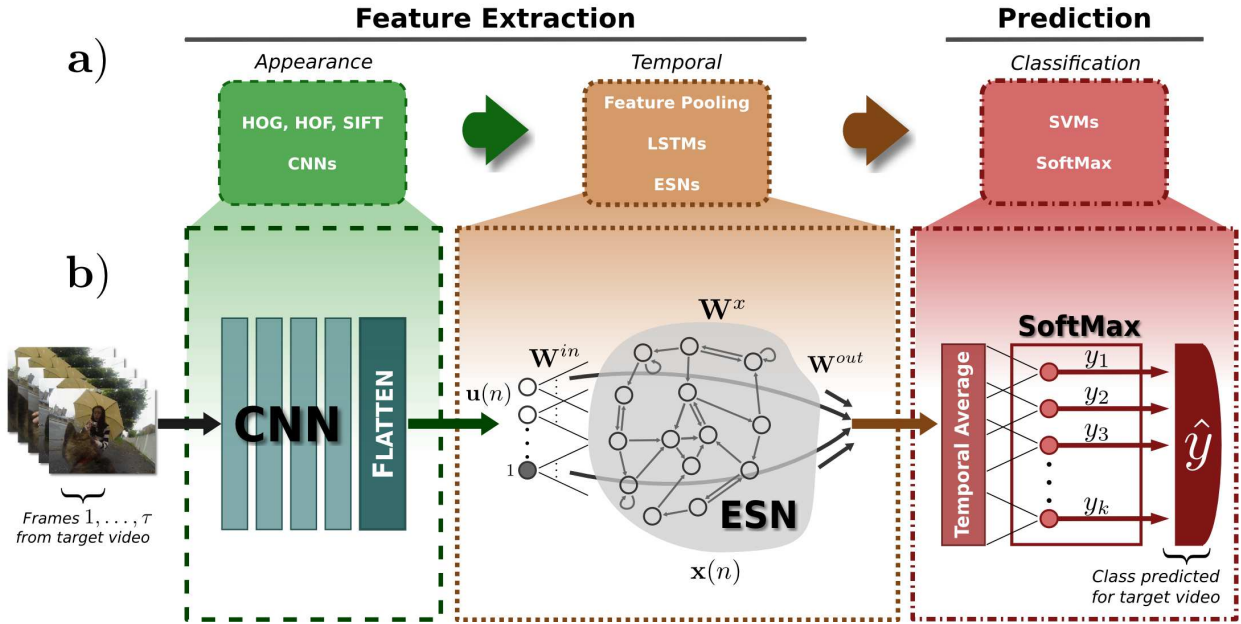}
	\caption{High level overview of the CDN architecture. Each frame $n = 1, \dots , \tau$ from video $v$ are passed through the CNN feature extractor to produce $\mathbf{u(n)}=\mathbf{u}^{(v)}_{n}$. All ESN responses $\mathbf{U}^{(v)}$ are collected (i.e. $\mathbf{U}^{(v)} = \{ \mathbf{u}^{(v)}_{1}, \dots , \mathbf{u}^{(v)}_{\tau} \} = \mathbf{x}(n) \ \text{for} \ n=1,\dots,\tau $). Temporal averaging is performed on $\mathbf{U}^{(v)}$. Finally the averaged responses are passed into the SoftMax layer for  video activity class prediction $\mathbf{y}(v)$.}
    \label{fig:networks}
    \end{center}
\end{figure*}

\section{Convolutional Drift Neural Network}
In this paper we propose a neural network architecture that requires minimal training to achieve competitive performance on spatio-temporal tasks (e.g. video classification). We demonstrate our architecture concept by designing an example network to perform video-level activity classification on egocentric videos. Our experimental networks are designed to perform this task specifically, but the concept is general enough that it should be readily adaptable for similar tasks. At a conceptual level, our proposed architecture contains: a CNN appearance feature extractor, an ESN temporal feature extractor, temporal information aggregation, and a predictor. 


In this initial investigation, we are primarily motivated to find task performance while minimizing training time and resources. This section describes an example network for video-level activity classification, with a focus on minimal training and hyper-parameter tuning. In the example network, CNN feature extractors are frozen to prevent training, and ESN reservoir weights are inherently fixed. This leaves the classification layer as the only place learning can occur. Essentially, we are attempting to achieve competitive performance on a difficult spatio-temporal task by training only a \emph{single} neural network layer. An overview of our proposed example network is shown in Fig. \ref{fig:networks}.

\subsection{CNN Feature Extractors} \label{sec:arch_FE}
Features produced by a CNN trained on natural images tend to be generalizable \cite{Yosinski2014}, so they can often be applied to new tasks with little or no additional training. Most publicly available pre-trained CNNs are high-performing models, trained on a large natural image dataset (e.g. ImageNet). Transfer learning allows us to leverage the power of these models for video analysis when using them as feature extractors. Feature extractors were constructed using the typical transfer learning approach, where part of an existing pre-trained CNN model is used. We identify a layer in the source CNN to use as a feature extraction point. All layers beyond that point are then removed, leaving behind a feature extractor network.

Feature extractors are applied to video data by considering each video as a collection of frame images. At the frame level, feature extraction abstracts raw, high-dimensional visual information as a per-frame feature vector, $\mathbf{u}^{(v)}_{n} $, shown in Eq. \ref{equ:arc_FE_framevec}.
\begin{align}
\mathbf{u}^{(v)}_{n} &= ( u_{1}, u_{2}, \dots , u_{m} ) \label{equ:arc_FE_framevec}
\end{align}
where, $n$ is the frame sequence position, $m$ is the output dimension of a DCNN feature extractor, and $v$ is a unique video identifier.

\noindent
At the video level, features are represented as the set of all $\mathbf{u}^{(v)}_{n}$ frame feature vectors used for prediction, $\mathbf{U}^{(v)}$, shown in Eq. \ref{equ:arc_FE_videovec}.
\begin{align}
\mathbf{U}^{(v)} &= \{ \mathbf{u}^{(v)}_{1}, \dots , \mathbf{u}^{(v)}_{\tau} \} \label{equ:arc_FE_videovec}
\end{align}
where $\tau$ is the number of frames evaluated in video $v$. 




\subsection{ESN, Temporal Averaging, \& Prediction}
Temporal features are extracted with an ESN that has been slightly modified from the base model to simplify computation. Since our experimental tasks require classification, temporal information is aggregated using a simple feature-wise temporal averaging of reservoir activations over all frames in a video. Finally, classification is performed on the fused representation using a single SoftMax classifier layer which predicts video activity labels. 

Our ESN implementation differs from the basic model described in Sec. \ref{sec:bg_esn} in a few key areas. During preliminary testing, we found that a rectified linear unit (ReLu) activation for reservoir neurons actually out-performed the hyperbolic tangent. ReLu activations are also very simple to calculate. This motivated us to use ReLu neurons in the reservoir, a non-traditional approach for ESNs. ReLu activation also greatly simplifies our ESN activation and update functions, as shown in Eqn. \ref{equ:esn_relu_act_1} and Eqn. \ref{equ:esn_relu_act_2}. ESN response $\mathbf{x}(n)$ is computed for each frame in a video sample.

\begin{equation}
  \mathbf{\tilde{x}}(n) =
  \begin{cases}
    0 & \text{, $n < 0$} \\
	\mathbf{W}^{in} [1;\mathbf{u}(n)] + \mathbf{W}^{x}\mathbf{x}(n-1) & \text{, $n \geq 0$} \\
  \end{cases}
  \label{equ:esn_relu_act_1}
\end{equation}
\begin{equation}
  \mathbf{x}(n) =
  \begin{cases}
    0 & \text{, $n < 0$} \\
	( 1 - \alpha )  \mathbf{x}(n-1) + \alpha \mathbf{\tilde{x}}(n) & \text{, $n \geq 0$} \\
  \end{cases}
  \label{equ:esn_relu_act_2}
\end{equation}

There are several potential methods for collecting and interpreting reservoir neuron responses \cite{Lukosevicius2012}. Response aggregation techniques using concatenation or pooling are common choices to fuse temporal information for prediction. Given the classification task in our experiments, we use the simple time-averaging technique shown in Eqn. \ref{equ:esn_basic_clf_4}, to compute $\Sigma \mathbf{y}$.

Time-averaged reservoir responses for each video sample are classified by a single SoftMax classification layer. This layer is trained using standard neural network optimization methods to adjust ESN output weights $\mathbf{W}^{out}$, minimizing the categorical cross-entropy loss between  $\mathbf{y}(n)$ and $\mathbf{y}^{target}(n)$.


\section{Experiments}
\subsection{Datasets}
Experiments were performed with two first-person video datasets: the DogCentric activity dataset \cite{Iwashita2014} and the UEC\-Park dataset \cite{Kitani2011}. Both datasets were recorded using wearable cameras. 

\begin{figure}[tb]
	\begin{center}
		\includegraphics[width=1\linewidth]{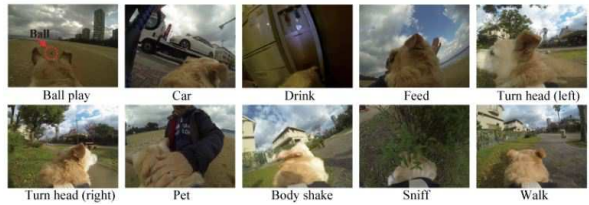}
	\end{center}
	\caption{Example Frames and Activity Classes from the Dogcentric Dataset. Figure and dataset by Iwashita et al. \cite{Iwashita2014}.}
	\label{fig:dogcentric_example}
\end{figure}

DogCentric contains 208 videos recorded by cameras attached to the backs of dogs while they performed 10 different classes of activity (e.g. playing with a ball, sniffing an object). Fig. \ref{fig:dogcentric_example} provides an example of frames extracted from DogCentric. The motion displayed throughout the video is often erratic and unpredictable, and therefore it made for a difficult video activity challenge. Videos in DogCentric do not have a fixed length. While an ESN with time-averaged outputs can be used to process variable sequence length data, we instead chose use a fixed sequence length of 160 frames for simplicity. Shorter videos received zero padding, and longer videos were truncated to provide a consistent 160 frames. This sequence length was determined by a simple inspection of DogCentric to find the mean frame count and distribution, then choosing a value that was likely to capture features for the majority of videos.

UEC\-Park is a dataset sourced from a single 25 minute long first-person video. It is recorded by a head-mounted camera worn by a human subject performing sports-related activities in a park (e.g. jogging, twisting, resting). The UEC-Park video was separated into 2 second segments, producing 766 unique equal length video sequences. There are 29 total activity classes, with segments labeled by the most dominant activity present. Since UEC-Park has fixed-length segments, no sequence padding or truncation was used.

\subsection{CNN Feature Extractor Implementation}
We considered two CNN models as feature extractor networks in this study, VGG-16 and ResNet-50. 
The VGG-16 feature extractor was created by truncating the source network after the first fully connected layer in the final block. Truncation of the ResNet-50 network was applied after the final average pooling layer. The size of feature vectors produced by each are 4096 and 2048, respectively. The ResNet-50 extractor output for each frame is multi-dimensional, so it is flattened to produce one-dimensional feature vector.

\subsection{ESN Implementation}
ReLu reservoir activations simplify ESN computation along with hyper-parameter tweaking to further reduce complexity. The leakage term in the reservoir is not applied (i.e. $\alpha = 1$). Without leakage, activation and update equations are equivalent(i.e. $\mathbf{\tilde{x}}(n) \equiv \mathbf{x}(n)$). We also do not scale the spectral radius $\rho$, allowing it to remain at the initialization value equal to the maximal absolute eigenvalue of $\mathbf{W^x}$. In practice, both of these parameters are usually tuned for performance, but in this initial baseline exploration we chose to keep the values fixed for simplicity.

\subsection{Classification}
Activities in each dataset were classified using the SoftMax layer with a number of neurons equal to the number of classes present (i.e. $N_y=k$). In the DogCentric dataset, this made $N_y = 10$, and in UEC-Park $N_y = 29$. Training was executed using Adam optimization \cite{kingma2014adam}, as we found it converged in far fewer epochs than gradient descent during preliminary testing.

\subsection{Experimental Setup \& Evaluation}
Experiments were performed using half of the videos per activity class placed in the training set, and half in the testing set. On uneven splits, the extra video was placed in the testing set. Random training/testing splits were repeated 100 times with a stratified random shuffle. In each experiment several network configurations are selected, and trained over 1600 epochs for all train/test splits. Performance was evaluated by averaging test set classification accuracy results for a given network configuration over all data splits. 

\subsection{Experiments on DogCentric}
\begin{figure}[tb]
	\begin{center}
		\includegraphics[width=.9\linewidth]{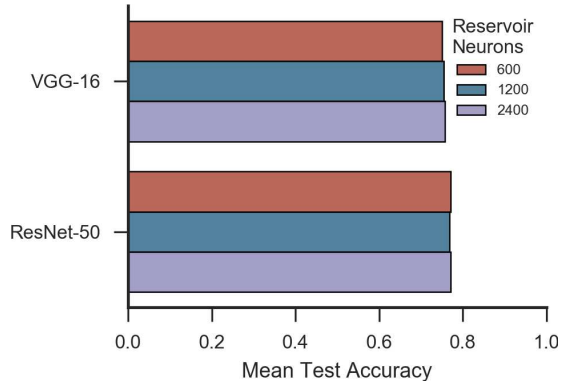}
	\end{center}
	\caption{Mean test accuracy on the DogCentric dataset. Networks trained over 1600 epochs with 600, 1200, and 2400 reservoir neurons (ReLu activations). Performance averaged over 100 random data split replications.}
	\label{fig:dogcentric_experiment_fe-compare_bar}
\end{figure}

Six different network configurations were evaluated on the DogCentric dataset by testing both the VGG-16 and ResNet-50 feature extractors, and varying the number of reservoir neurons at 600, 1200, and 2400. Fig. \ref{fig:dogcentric_experiment_fe-compare_bar} shows a summary comparison of performance, with details provided in Tbl. \ref{tab:dogcentric_experiment_results}. 

\begin{figure}[tb]
	\begin{center}
		\includegraphics[width=.9\linewidth]{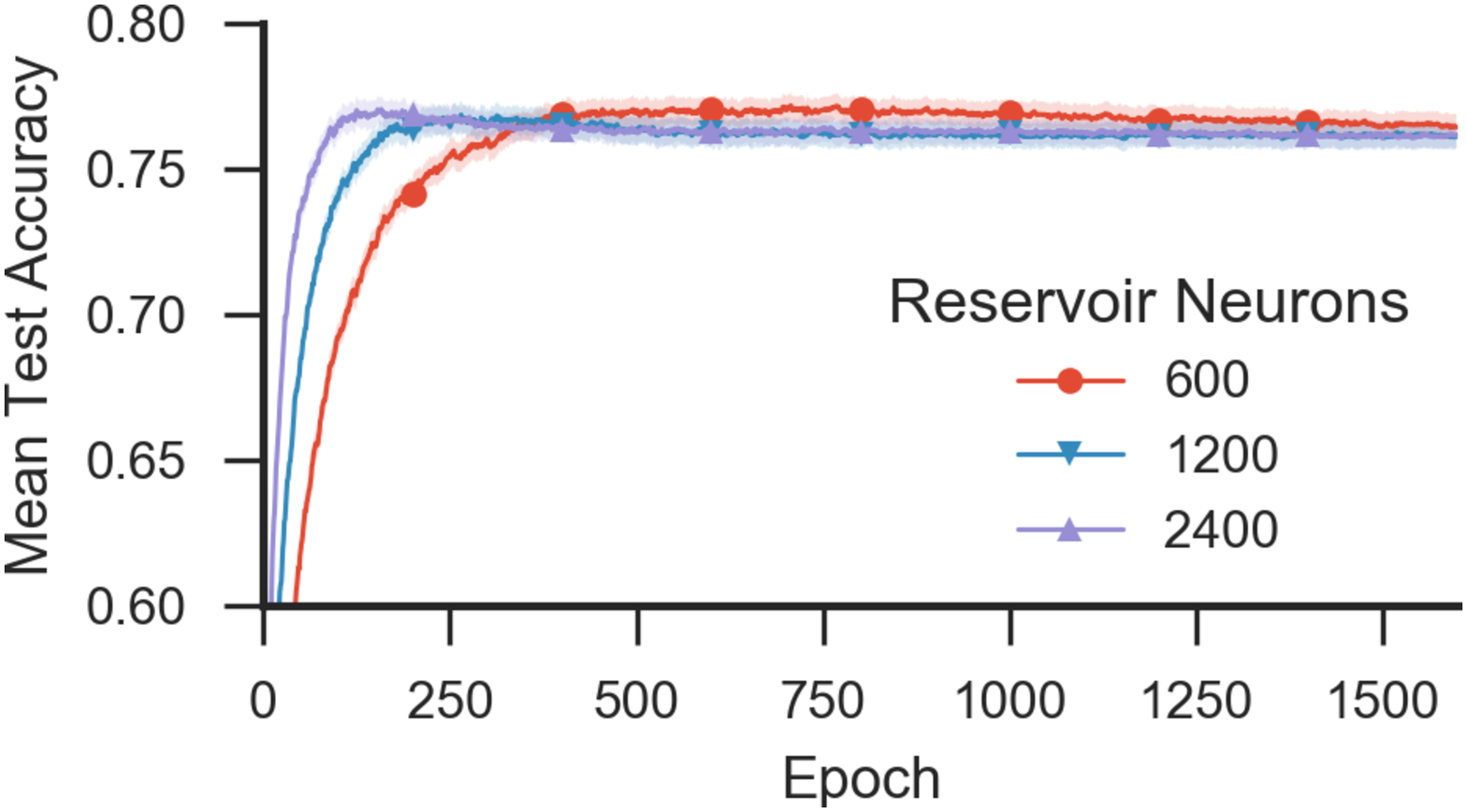}
	\end{center}
	\caption{Mean test accuracy on the DogCentric dataset. Networks trained over 1600 epochs with 600, 1200, and 2400 reservoir neurons (ReLu activations). Performance averaged over 100 random data split replications. Y-axis zoomed for readability.}
	\label{fig:dogcentric_experiment_resnet_line}
\end{figure}

We observed that the feature extractor network model consistently shows an impact on performance, with the ResNet-50 feature extractor outperforming VGG-16 in every test. Reservoir neuron count showed less of an impact on achieving the best accuracy. Testing with different reservoir sizes  produced consistently similar scores when using the same feature extractor. However, networks with more reservoir neurons did tend to attain this level of accuracy in fewer training epochs, as shown in Fig. \ref{fig:dogcentric_experiment_resnet_line}. The ResNet-50 configuration with 600 reservoir neurons achieved the best overall accuracy numerically at 77.2\%, but all results were close enough to be considered approximately equivalent.

\begin{table}
	\centering
	\caption{Experiment results on the DogCentric dataset. }\label{tab:dogcentric_experiment_results}
	\begin{tabularx}{.95\columnwidth}{X rcc >{\arraybackslash}X}
	& \multicolumn{3}{c}{\textbf{Best Mean Test Accuracy: DogCentric}} & \\
	\cmidrule(lr){2-4}
	& \multirow{2}{*}{\shortstack{Reservoir\\Neurons}} & \multirow{2}{*}{ResNet-50} & \multirow{2}{*}{VGG-16} & \\ \\
	\cmidrule(lr){2-2} \cmidrule(lr){3-3} \cmidrule(lr){4-4} 
	& 600     &    \textbf{77.2 \%} &    75.1 \% & \\
	& 1200    &    76.9 \% &    75.4 \% & \\
	& 2400    &    77.1 \% &    75.8 \% & \\
	\cmidrule(lr){2-4}
	& \multicolumn{3}{c}{\scriptsize{DogCentric Experiment: \textit{100 train/test splits, 1600 epochs}}} & \\
	\cmidrule(lr){2-4}
	\end{tabularx}	
\end{table}



\subsection{Experiments on UEC-Park}
\begin{figure}[tb]
	\begin{center}
		\includegraphics[width=.9\linewidth]{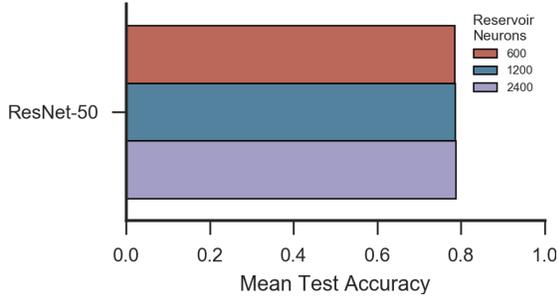}
	\end{center}
	\caption{Mean test accuracy on the UEC-Park dataset. Networks trained over 1600 epochs with 600, 1200, and 2400 reservoir neurons (ReLU activations). Accuracy averaged over 100 random data split replications.}
	\label{fig:uec-park_experiment_bar}
\end{figure}

Three different network configurations were evaluated on the UEC-Park dataset. ResNet-50 was shown to be a superior feature extractor across all configurations in both preliminary testing and the DogCentric experiments, so it was used on UEC-Park. Networks with 600, 1200, and 2400 reservoir neurons were tested. Fig. \ref{fig:uec-park_experiment_bar} shows a summary comparison of performance, with details provided in Tbl. \ref{tab:uec-park_experiment_results}. 

\begin{figure}[tb]
	\begin{center}
		\includegraphics[width=.9\linewidth]{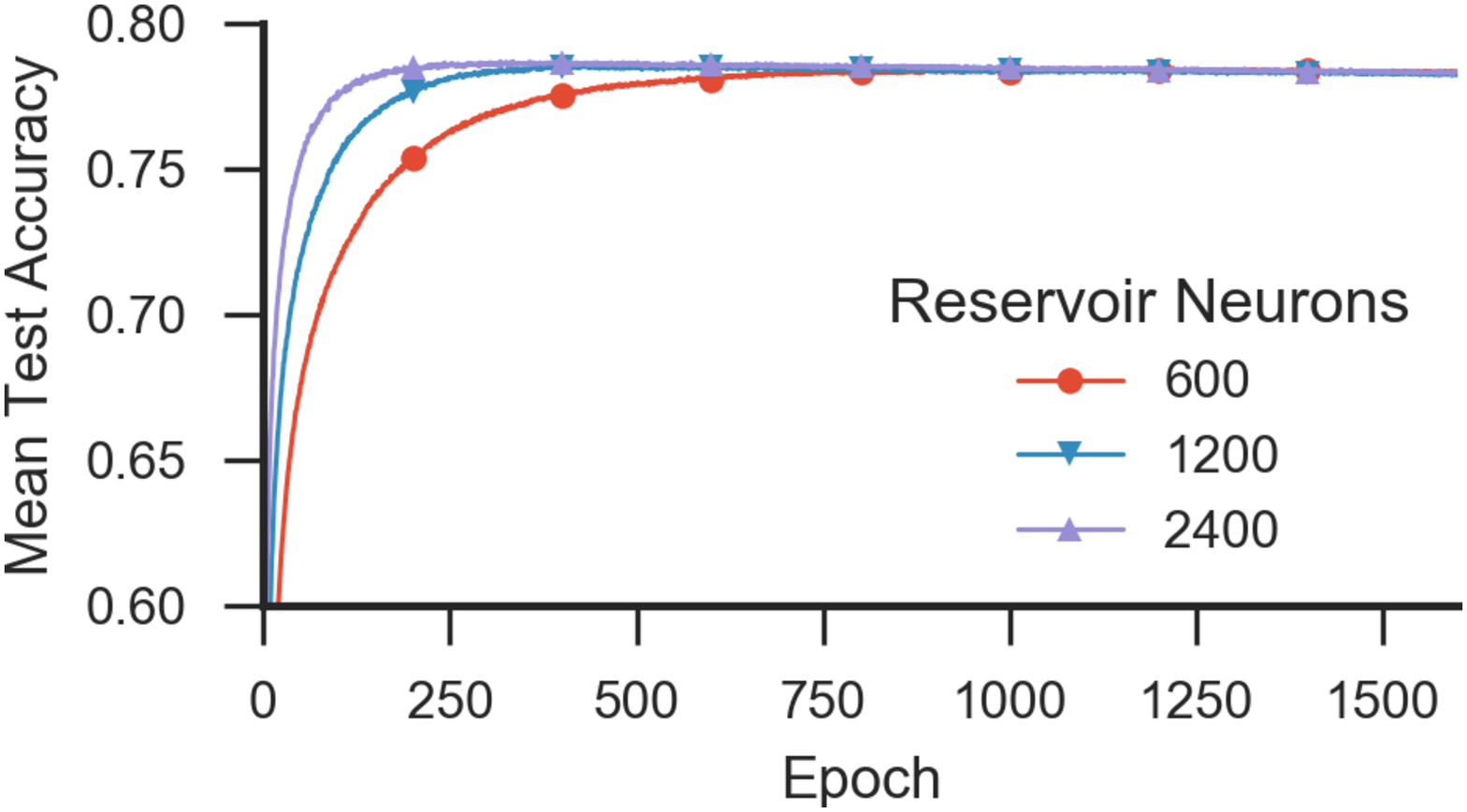}
	\end{center}
	\caption{Mean test accuracy on the UEC-Park dataset. Networks trained over 1600 epochs with 600, 1200, and 2400 reservoir neurons (ReLU activations). Accuracy averaged over 100 random data split replications. Y-axis zoomed for readability.}
	\label{fig:uec-park_experiment_resnet_line}
\end{figure}

As in the DogCentric experiments, configurations with different reservoir neuron counts produced very similar performance results. In Fig. \ref{fig:uec-park_experiment_resnet_line}, the mean accuracy plot demonstrates that again, more neurons seem to contribute to quicker convergence toward the best accuracy value observed.

\begin{table}
	\centering
	\caption{Experiment results on the UEC-Park dataset. }\label{tab:uec-park_experiment_results}
	\begin{tabularx}{.95\columnwidth}{X rc >{\arraybackslash}X}
		& \multicolumn{2}{c}{\textbf{Best Mean Test Accuracy: UEC-Park}} & \\
		\cmidrule(lr){2-3}
		& \multirow{2}{*}{\shortstack{Reservoir\\Neurons}} & \multirow{2}{*}{ResNet-50} &  \\ \\
		\cmidrule(lr){2-2} \cmidrule(lr){3-3} 
		& 600  &    		78.5 \%  & \\
		& 1200  &   		78.6 \%  & \\
		& 2400 &    \textbf{78.7 \%} & \\
		\cmidrule(lr){2-3}
		& \multicolumn{2}{c}{\scriptsize{UEC-Park Experiment: \textit{100 train/test splits, 1600 epochs}}} & \\
		\cmidrule(lr){2-3}
	\end{tabularx}	
\end{table}

\subsection{Comparison to State-of-the-Art}
\label{subsec:Result Comparison}
We compare our activity classification results with others from literature. Our interest was to develop an end-to-end trainable neural network for video level classification, which precluded the use of hand-crafted features (HCF). In the literature for egocentric video, this type of network is fairly uncommon (i.e. HCFs are usually used), and few examples suitable for direct comparison were available. Instead, we provide all results found on the two datasets, separately labeling methods that use HCF, and those that don't.

\subsubsection{DogCentric Comparison}
Tbl. \ref{tab:dogcentric_sota_compare} provides comparative results on the DogCentric dataset. The two best performance results in literature were both obtained using HCFs, and both with a TDD + Temporal Filters approach \cite{Piergiovanni2016a}. The top performer used an LSTM attention mechanism to achieve $81.4\%$. The other model does not use LSTM attention, achieving $79.6\%$ accuracy. Our best result of $77.2\%$ with a ResNet-50 CDN did not exceed these two results, but it did perform better than all other HCF-based methods. 

\begin{table}
	\centering
	\caption{Comparison of state-of-the-art accuracy results on the DogCentric dataset. 10 activities, classification accuracy over 100 data splits.}\label{tab:dogcentric_sota_compare}
	\begin{tabularx}{.95\columnwidth}{clr >{ \arraybackslash}X}		
 &	\textbf{Approach}	&	\textbf{Accuracy}	&	\\
		\cmidrule(lr){1-3}
		\multirow{8}{*}{\scriptsize HCF }
 & HOG+HOF+LBP+Cub.+Opt.Fl.\cite{Iwashita2014}      & 60.5\%           & \\
 & ITF\cite{Piergiovanni2016a,wangtrajectory}                		& 67.7\%           & \\
 & ITF+CNN\cite{Piergiovanni2016a,UniversityChallenge}       		& 69.2\%           & \\
 & POT\cite{Ryoo2015}                                   		& 73.0\%           & \\
 & POT+ITF\cite{Ryoo2015}                               		& 74.5\%           & \\
 & TDD\cite{wangaction2015,Piergiovanni2016a}                		& 76.6\%           & \\
 & TDD+Temp. Fil. \cite{Piergiovanni2016a}             		& 79.6\%           & \\
 & TDD+Temp. Fil.+LSTM\cite{Piergiovanni2016a}       		& 81.4\%           & \\
 \cmidrule(lr){1-3}
 \multirow{7}{*}{\scriptsize No HCF}
 & VGG+Max Pooling\cite{Piergiovanni2016a}                			& $\approx$ 57.2\% & \\
 & VGG+Mean Pooling\cite{Piergiovanni2016a}                		& 59.9\%           & \\
 & VGG+Sum Pooling\cite{Piergiovanni2016a}                 		& 59.9\%           & \\
 & VGG+Temp. Fil.-Learned\cite{Piergiovanni2016a}      		& $\approx$ 65.0\% & \\
 & VGG+Temp. Fil.-Learned+LSTM\cite{Piergiovanni2016a} 	& $\approx$ 65.0\% & \\
 & \textbf{CDN (VGG-16)}                                		& \textbf{75.8 \%} & \\
 & \textbf{CDN (ResNet-50)}                             		& \textbf{77.2 \%} & \\
		\cmidrule(lr){1-3} 
		\multicolumn{4}{c}{\scriptsize{(\textbf{Bold}: Our approach) ($\approx$: est. from graph)}}\\
		\cmidrule(lr){1-3}
	\end{tabularx}	
\end{table}

Among non-HCF methods, our ResNet-50 CDN model did achieve the best performance of $77.2\%$. Other models in this category were developed by Piergiovanni et al., and they all use a VGG feature extractor along with various pooling or temporal filtering techniques. One of these models offers the most direct comparison to our approach, since it is somewhat similar (i.e. it uses a CNN and an RNN). That model contains a VGG feature extractor, learned temporal filters, LSTM attention mechanism, and a SoftMax classifier. In a direct comparison of VGG-based methods, our VGG-16 CDN achieved an accuracy score $\approx 10.8\%$ higher. Given the model similarities for non-temporal elements, this result provides some insight into the relative contribution of an ESN for the video-level activity classification task.

\subsubsection{UEC-Park Dataset Comparison}
In literature, all results found on the UEC-Park dataset were produced using HCFs, so non-HCF methods were not available for direct comparison. As shown in Tbl. \ref{tab:uec-park_sota_compare}, our ResNet-50 CDN model obtained a $78.7\%$ accuracy. Our approach performed slightly worse than the two HCF methods using POT \cite{Ryoo2015}, with a difference of $0.8\%$.

\begin{table}
	\centering
	\caption{Comparison of state-of-the-art accuracy results on the UEC-Park dataset. 29 activities, classification accuracy over 100 data splits.}\label{tab:uec-park_sota_compare}
\begin{tabularx}{.95\columnwidth}{cXlr >{ \arraybackslash}X}		
 & &	\textbf{Approach}	&	\textbf{Accuracy}	&	\\
 \cmidrule(lr){1-4}
 \multirow{7}{*}{\scriptsize HCF} 
 & & STIP+IFV \cite{laptev2005,Ryoo2015}    & 69.1\% &           \\
 & & Cubiod+IFV \cite{dollar2005,Ryoo2015}  & 72.3\% &           \\
 & & ITF+CNN \cite{UniversityChallenge}     & 75.7\% &           \\
 & & IFV+Pooling \cite{Ryoo2015}            & 76.4\% &           \\
 & & BoW+Pooling \cite{Ryoo2015}            & 76.5\% &           \\
 & & Inria ITF+IFV \cite{wang2013,Ryoo2015} & 76.6\% &           \\
 & & POT \cite{Ryoo2015}                    & 79.4\% &           \\
 & & POT+ITF \cite{Ryoo2015}                & 79.5\% &           \\
 \cmidrule(lr){1-4}
 \multirow{1}{*}{\scriptsize No HCF}
 & & \textbf{CDN (ResNet-50)}                   & \textbf{78.7\%} &  \\
\cmidrule(lr){1-4}
\multicolumn{4}{c}{\scriptsize{(\textbf{Bold}: Our approach)}}\\
\cmidrule(lr){1-4}
	\end{tabularx}	
\end{table}

\section{Conclusions}
We introduced a new method for combining CNNs and ESNs into a neural network architecture capable of performing complex spatio-temporal tasks with very little training or tuning. Our method was demonstrated to effectively process frame-level CNN features into video-level predictions on two different egocentric video datasets, producing accuracy results comparable to all state-of-the art approaches. Unlike most previous work on this video analysis task, we use no hand-crafted features, and our architecture is trainable end-to-end when desired. This result motivates us to explore this architecture further in future work by building on our baseline model. 

\ifCLASSOPTIONcaptionsoff
  \newpage
\fi



\bibliographystyle{IEEEtran}
\bibliography{dcnn-rc,added_bibs,local_bib}

\end{document}